\documentclass[10pt,twocolumn,letterpaper]{article}

\usepackage{cvpr}              

\usepackage{bm}

\definecolor{cvprblue}{rgb}{0.21,0.49,0.74}
\usepackage[pagebackref,breaklinks,colorlinks,allcolors=cvprblue]{hyperref}

\def\paperID{9613} 
\def\confName{CVPR}
\def\confYear{2025}

\title{High-Fidelity Relightable Monocular Portrait Animation with Lighting-Controllable Video Diffusion Model}

\author{
	Mingtao Guo\textsuperscript{1}\hspace{3em} Guanyu Xing\textsuperscript{1}\hspace{3em} Yanli Liu\textsuperscript{12}$^{\dag}$ \\
	{\textsuperscript{1}National Key Laboratory of Fundamental Science on Synthetic Vision, } \\
	{Sichuan University, Chengdu, China, 610065} \\
	{\textsuperscript{2}College of Computer Science, Sichuan University, Chengdu, China, 610065} \\
	{\tt\small mingtaoguo@stu.scu.edu.cn,\{xingguanyu, yanliliu\}@scu.edu.cn}
}

\begin{document}
\renewcommand{\thefootnote}{\fnsymbol{footnote}}
\twocolumn[{
\maketitle
\begin{center}
    \captionsetup{type=figure}
    \includegraphics[width=0.9\linewidth]{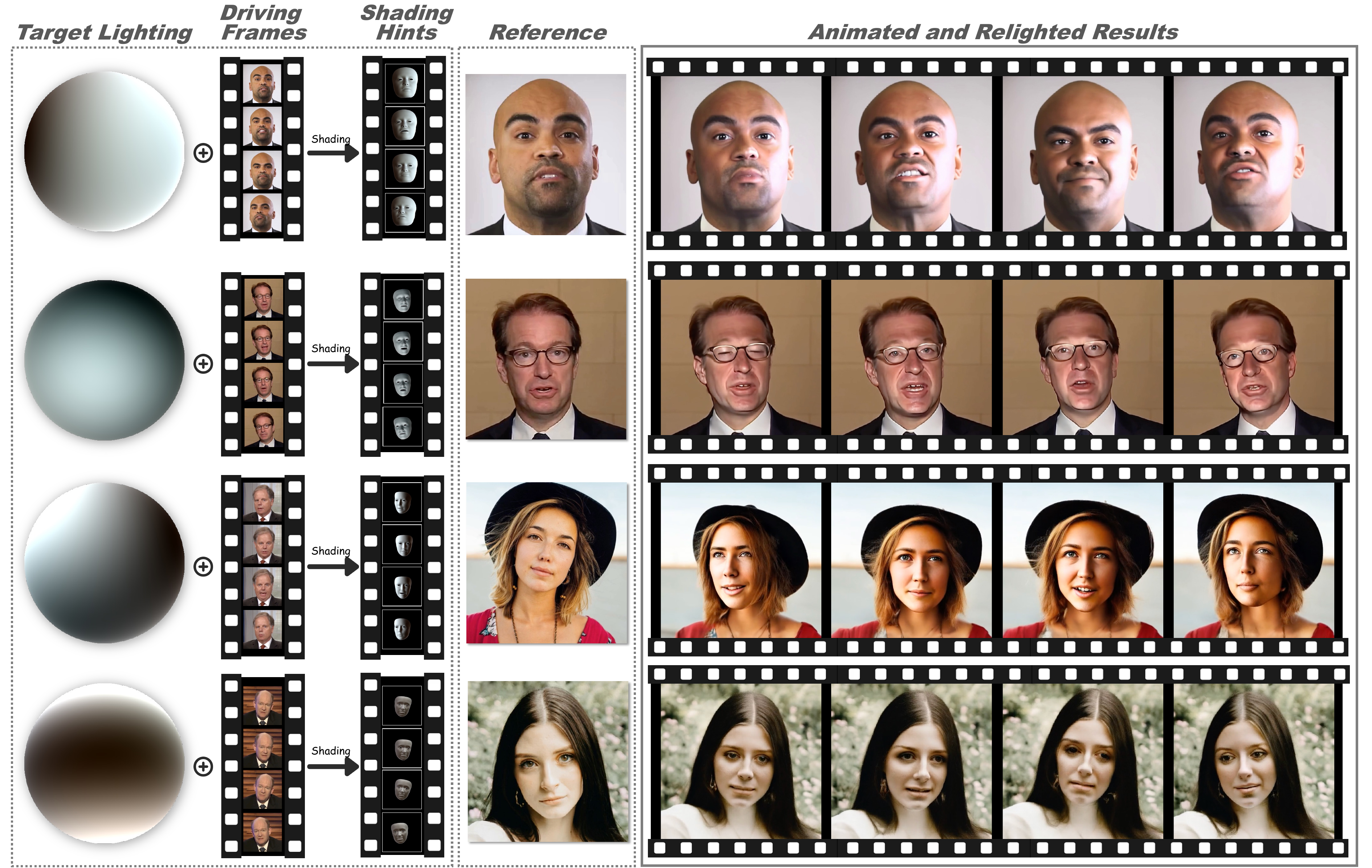}
    \captionof{figure}{Qualitative results of our method. The target lighting is applied to the meshes of the driving frames to generate shading hints. Using the shading hints, our relightable portrait animation framework animates and relights the reference frame, e.g., the results within the solid boxes show lighting consistent with the target lighting and poses consistent with the driving frames.}
    \label{fig:teaser}
\end{center}
}]
\footnotetext{\textsuperscript{\dag}Corresponding author: Yanli Liu.}
\begin{abstract}
Relightable portrait animation aims to animate a static reference portrait to match the head movements and expressions of a driving video while adapting to user-specified or reference lighting conditions. Existing portrait animation methods fail to achieve relightable portraits because they do not separate and manipulate intrinsic (identity and appearance) and extrinsic (pose and lighting) features. In this paper, we present a Lighting Controllable Video Diffusion model (LCVD) for high-fidelity, relightable portrait animation. We address this limitation by distinguishing these feature types through dedicated subspaces within the feature space of a pre-trained image-to-video diffusion model. Specifically, we employ the 3D mesh, pose, and lighting-rendered shading hints of the portrait to represent the extrinsic attributes, while the reference represents the intrinsic attributes. In the training phase, we employ a reference adapter to map the reference into the intrinsic feature subspace and a shading adapter to map the shading hints into the extrinsic feature subspace. By merging features from these subspaces, the model achieves nuanced control over lighting, pose, and expression in generated animations. Extensive evaluations show that LCVD outperforms state-of-the-art methods in lighting realism, image quality, and video consistency, setting a new benchmark in relightable portrait animation.
\end{abstract}
    
\section{Introduction}
\label{sec:intro}
Portrait animation has wide applications in video conferencing, virtual reality, and the film industry. With the rapid advancement of GANs \cite{goodfellow2014generative, karras2019style,karras2020analyzing} and Diffusion Models \cite{ho2020denoising,song2020denoising,rombach2022high}, existing portrait animation methods \cite{zhao2022thin, zhang2023sadtalker, guo2024liveportrait, xie2024x, chen2024echomimic} have demonstrated remarkable capabilities in generating talking faces. For instance, the state-of-the-art LivePortrait \cite{guo2024liveportrait} achieves real-time, high-fidelity portrait animation by designing better motion transformation and scaling up the talking head datasets. However, the ability of manipulating lighting during portrait animation remains under-explored, which is highly important for seamlessly blending the generated foreground portrait with backgrounds under varying lighting conditions.

In this paper, we focus on relightable portrait animation. We aim to animate a portrait in a still reference image, matching the head movement and facial expression of a driving video, at the same time, matching the lighting condition provided by users or extracted from another given portrait image. From the perspective of face attributes, we can reduce the task to preserving the intrinsic features (identity and appearance) of the reference portrait while effectively transferring the extrinsic ones (given pose and lighting) to the reference portrait. Obviously, the exact separation between intrinsic features and extrinsic features is crucial to reach our goal. A main reason why existing portrait animation methods can’t manipulate lighting is that they can't separately manipulate these two kinds of features.

In order to achieve high-fidelity relightable portrait animation, our key idea is to distinguish these two types of features by learning their feature subspaces, then maintain the intrinsic facial features and transfer external features. We observe that the image-to-video (I2V) diffusion model \cite{blattmann2023stable}, trained on a large-scale dataset encompassing a variety of portraits with different lighting, poses, identities and appearances, provides a foundation for learning the two feature subspaces. Specifically, we represent the portrait’s extrinsic attributes using shading hints rendered with the reference image’s 3D mesh, target lighting, and pose, while the intrinsic attributes are represented by the reference image. During the self-supervised training process, we design a shading adapter to map the shading hints into the extrinsic feature subspace and a reference adapter to map the reference image into the intrinsic feature subspace. By merging the features from these two subspaces, the model generates portraits with specified lighting, pose, identity, and appearance.

With the I2V diffusion model, we propose a novel \textbf{L}ighting \textbf{C}ontrollable \textbf{V}ideo \textbf{D}iffusion model (\textbf{LCVD}) to achieve high-fidelity, relightable portrait animation. First, we use an off-the-shelf model \cite{feng2021learning} to extract the 3D mesh, pose, and spherical harmonics lighting coefficients of the target portrait, which are rendered into shading hints containing lighting and pose information. In the training stage, to enable pose alignment and lighting control, we use a shading adapter to map these shading hints to the extrinsic feature subspace, representing external portrait attributes by establishing a mapping between the shading hints and the target portrait. For identity and appearance preservation, we use a reference adapter to map the reference image to the intrinsic feature subspace, representing internal portrait attributes by creating a mapping between an initial frame and subsequent frames. Finally, during the inference stage, we merge the features from the two subspaces and input them into the I2V diffusion model to generate portraits with specified lighting, pose, identity, and appearance. To further control the lighting magnitude, we employ multi-condition classifier-free guidance to emphasize the influence of the shading adapter and reduces the reference’s impact on relighting.

Our main contributions are listed as follows:
\begin{itemize}
\item We introduce the Lighting Controllable Video Diffusion model, a diffusion-based framework for relightable portrait animation, which overcomes the limitations of current portrait animation methods that fail to manipulate lighting while animate the portrait.
\item We propose a shading adapter and a reference adapter to construct feature subspaces for extrinsic and intrinsic facial features. By merging these two subspaces, the I2V model is guided to achieve relightable portrait animation.
\item Extensive experiments demonstrate that LCVD surpasses state-of-the-art methods, showing significant improvements in metrics related to lighting effects, image quality, and video consistency.
\end{itemize}

\section{Related Work}
\label{sec:relatedwork}
\subsection{Portrait Relighting}
Portrait relighting involves adjusting the lighting of an image or video while preserving the subject’s identity and appearance. Previous methods \cite{pandey2021total, sun2019single, zhang2021neural} utilized One-Light-at-a-Time (OLAT) systems to capture detailed geometry and reflectance, achieving realistic relighting. However, OLAT data is expensive and difficult to acquire, limiting its practicality. To overcome this, recent approaches \cite{zhou2019deep, hou2021towards} simulate multi-lighting data and train networks for relighting. Despite these efforts, simulated methods still lag behind the realism achieved by OLAT-based techniques. Additionally, learning 3D face representations from 2D images without explicit 3D supervision has now become feasible. Recent method \cite{nerffacelighting} combines neural radiance fields (NeRF) \cite{mildenhall2021nerf} with generative models like GANs \cite{karras2020analyzing} and diffusion models \cite{rombach2022high} to generate high-resolution, multi-view consistent face images. 

Another simplified strategy \cite{nestmeyer2020learning, qiu2024relitalk} involves capturing a selfie video or a sequence of images to obtain multi-view information. However, the rendering quality is highly dependent on the accuracy of the geometry, requiring sufficient viewpoints from the video. Additionally, these methods often need to be retrained for each new video, which makes them impractical. In contrast, our method achieves high-fidelity and temporally stable video portrait relighting, requiring only a single portrait image and a target lighting.

\subsection{Diffusion-based Portrait Animation}
Denoising Diffusion Models \cite{ho2020denoising,song2020denoising} are based on the idea of Markov diffusion and fits the distribution of real samples by approximating a standard normal distribution. They outperform GANs \cite{goodfellow2014generative} in sample diversity and quality, and have been successfully applied to image synthesis \cite{kawar2023imagic, avrahami2023spatext, zhang2023adding, kumari2023multi, ruiz2023dreambooth}, image editing \cite{avrahami2023blended, brooks2023instructpix2pix}, and video synthesis \cite{yang2024cogvideox, chen2024videocrafter2, blattmann2023stable}. In portrait animation, FADM \cite{zeng2023face} refines coarsely animated portraits generated by previous methods \cite{siarohin2019first, siarohin2021motion, hong2022depth} by combining 3DMM \cite{blanz1999morphable} parameters with a diffusion model to improve appearance. Follow-Your-Emoji \cite{ma2024follow} instead uses expression-aware landmarks within the Animate-Anyone framework \cite{hu2024animate} to guide the animation process.

However, while diffusion-based portrait animation methods effectively transfer poses from driving images to animate the reference image, they cannot simultaneously manipulate the lighting of the portrait in the reference image during the animation.
\section{Preliminaries}
The Latent Diffusion Model (LDM) \cite{rombach2022high} is designed to generate high-quality, diverse images based on text prompts. It performs the denoising process within the latent space of a Variational Autoencoder (VAE) \cite{esser2021taming}. During training, the input image $\mathbf{x}_0$ is first encoded into its latent representation $\mathbf{z}_0 = \mathcal{E}(\mathbf{x}_0)$, where $\mathcal{E}(\cdot)$ represents the frozen encoder. The resulting latent code $\mathbf{z}_0$ is then perturbed as follows:
\begin{equation}
	\mathbf{z}_{t}=\sqrt{\bar{\alpha}_t}\mathbf{z}_0+\sqrt{1-\bar{\alpha}_t}\bm{\epsilon},\bm{\epsilon}\in \mathcal{N}(0,\bf{\textit{I}}),
	\label{eq:diffusion}
\end{equation}
where \( \bar{\alpha}_t = \prod_{i=1}^{t} (1 - \beta_t) \) with \( \beta_t \) is the noise strength at step \( t \), and \( t \) is sampled uniformly from \( \{1, \dots, T\} \). This process is a Markov chain that adds Gaussian noise to the latent code $\mathbf{z}_0$. The denoising model \( \bm{\epsilon}_\theta \) learns the latent space distribution by optimizing the objective function using \( \mathbf{z}_t \) as input,
\begin{equation}
	\mathcal{L}_{LDM}=\mathbb{E}_{\mathbf{z}_0,\mathbf{c},\bm{\epsilon}\in \mathcal{N}(0,\mathbf{\textit{I}})}\left[ \|\bm{\epsilon}-\bm{\epsilon}_{\theta}(\mathbf{z}_t,t,\mathbf{c}) \|^2_2\right],
	\label{eq:ldm}
\end{equation}
where $\mathbf{c}$ represents the condition, which is the text embedding encoded by the CLIP \cite{radford2021learning} text encoder provided by the user.
\section{Methodology}
\label{sec:method}
\subsection{Overview}
\begin{figure*}[!htbp]
	\centering
	\includegraphics[width=1\textwidth]{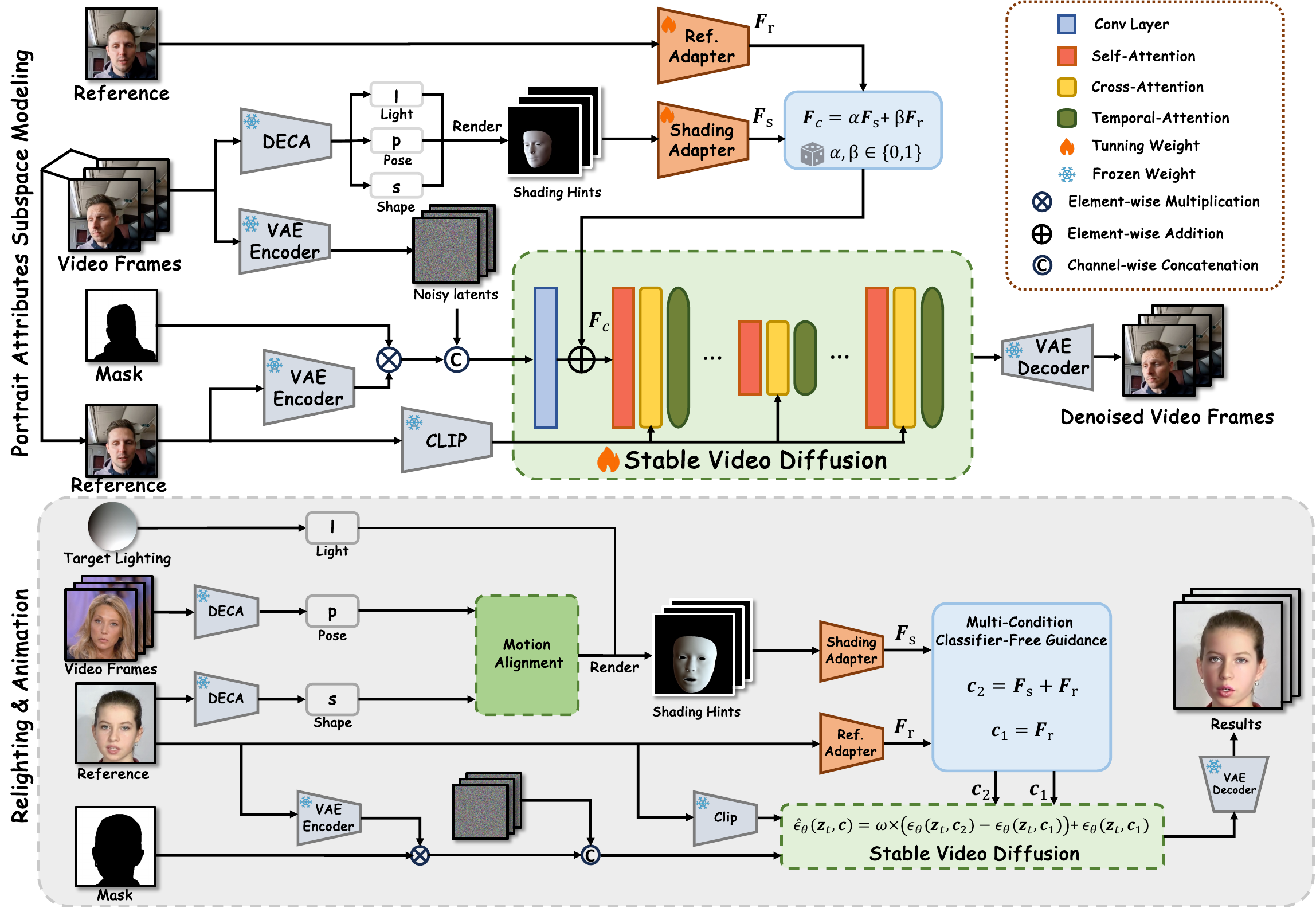}
	\caption{Overview of our pipeline for lighting controllable portrait animation. It consists of two main stages: (1) Portrait Attributes Subspace Modeling Stage: We use DECA to encode video frames and extract lighting, pose, and shape parameters, which are rendered as shading hints. After processing the shading hints and reference image through the shading adapter and reference adapter, the two features are randomly selected and fused as guidance to guide the Stable Video Diffusion Model in generating denoised video frames with consistent lighting, pose, identity, and appearance. (2) Relighting and Animation Stage: We render the shading hints using the pose of the portrait from the video, the shape from the reference image, and the spherical harmonics coefficients of the target lighting. After processing the shading hints and reference image through two adapters, we employ multi-condition classifier-free guidance to adjust the magnitude of the extrinsic feature guidance direction, enabling the generation of lighting controllable portrait animations.}
    \label{fig:overview}
    \vspace{-0.18in}
\end{figure*}

Our pipeline for lighting controllable portrait animation consists of two stages. First, in the training phase, we construct portrait intrinsic and extrinsic feature subspaces within a pre-trained I2V model’s feature space using two adapters. Then, in the relighting and animation stage, we modify the extrinsic subspace and merge it with the intrinsic subspace to achieve relightable portrait animation, as illustrated in Fig. \ref{fig:overview}.
\begin{enumerate}
    \item \textit{Portrait Attributes Subspace Modeling Stage:} We employ an off-the-shelf model DECA \cite{feng2021learning} to encode each frame of the input video, extracting key parameters such as lighting, pose, and shape, which are then rendered as shading hints. After the shading hints and reference image are processed through the shading adapter and reference adapter, they are randomly selected, with each training iteration potentially including either one, both, or neither for composition (Sec. \ref{sec:facm}). The composed features are subsequently fed into the Stable Video Diffusion Model \cite{blattmann2023stable} for self-supervised training (Sec. \ref{sec:agvdm}). The goal of this stage is to model both the extrinsic and intrinsic feature subspaces through the joint optimization of two adapters.
    \item \textit{Relighting and Animation Stage:} We render the shading hints using the pose of the portrait from the video, the shape from the reference image, and the spherical harmonics coefficients of the target lighting. Then, we combine the outputs of the shading adapter and the reference adapter to form the conditional set and employ multi-condition classifier-free guidance to adjust the magnitude of the extrinsic feature guidance direction by modifying the strength of the guidance, thereby generating results for lighting controllable portrait animation (Sec. \ref{sec:lcpa}).
\end{enumerate}

\subsection{Portrait Attributes Subspace Modeling}
\label{sec:facm}
Current portrait animation methods can be driven by user-provided pose information. However, since portrait intrinsic and extrinsic features are entangled during the self-supervised training process, manipulating lighting requires modifying extrinsic features. This entanglement makes it difficult to adjust lighting independently during portrait animation. Therefore, separating portrait intrinsic and extrinsic features is a significant challenge for enabling effective lighting control in portrait animation. 

To address this, we design a shading adapter and a reference adapter to construct an extrinsic feature subspace, as well as an intrinsic feature subspace within the SVD feature space during the training phase. First, we use the parametric model $\mathbf{FLAME}$ \cite{li2017learning} as a prior to model the shape and pose attributes of human portraits.
\begin{equation}
	\mathbf{FLAME}(s,p,e)=\mathbb{R}^{|s|\times |p|\times |e|} \rightarrow \mathbb{R}^{m\times 3},
	\label{eq:flame}
\end{equation}
which takes shape coefficients $s\in \mathbb{R}^{|s|}$, pose $p\in \mathbb{R}^{|p|}$, and expression $e\in \mathbb{R}^{|e|}$ as inputs to generate the corresponding 3D face mesh. We use DECA to estimate these parameters, with the added benefit of DECA’s ability to predict second-order spherical harmonics lighting coefficients $l\in \mathbb{R}^{|l|}$. We then render the 3D face mesh using the spherical harmonics to produce a lighting-shaded face, referred to as shading hints. 

We process the input video fragment into a sequence of shading hints, which, together with the reference image, are independently transformed by the shading and reference adapters into features $\mathbf{F}_s$ and $\mathbf{F}_r$. To establish these two feature subspaces, $\mathbf{F}_s$ and $\mathbf{F}_r$ are recombined with random coefficients $\left\{\alpha, \beta|\alpha,\beta\in \left\{0,1\right\}\right\}$ and input into the SVD. This method effectively enables the SVD to explore both the extrinsic and intrinsic feature subspaces within its feature space.

\subsection{Lighting-Guided Video Diffusion Model}
\label{sec:agvdm}
We choose the Stable Video Diffusion Model (SVD) \cite{blattmann2023stable} as the prior model for our LCVD method. However, SVD is an image-guided video generation model that takes an input image $I \in \mathbb{R}^{H \times W \times 3}$. This image is first encoded by CLIP’s vision encoder \cite{radford2021learning} and passed into SVD’s cross-attention module. At the same time, the image is encoded by a Variational Autoencoder (VAE) \cite{esser2021taming} into a latent representation $\mathbf{z}_0 = \mathcal{E}(I) \in \mathbb{R}^{h \times w \times c}$. This latent $\mathbf{z}_0$ is then replicated $T$ times and concatenated along the channel dimension with noise $\mathbf{\hat{z}} \in \mathbb{R}^{T \times h \times w \times c}$, resulting in $\mathbf{z}_t \in \mathbb{R}^{T \times h \times w \times 2c}$. The resulting $\mathbf{z}_t$ is then input into a 3D UNet \cite{ronneberger2015u}, which progressively denoises the input to generate a video of $T$ frames. Here, we set $h = H/8$, $w = W/8$, and $c = 4$.

For instance, when an image of a dog walking on the street is input into the SVD, the model predicts the next $T$ frames of the dog based on the original image. As a result, the subsequent frames generated by SVD inherit the objects and lighting conditions from the original image. To eliminate the influence of the original image’s lighting on the relighting results, as shown in Fig. \ref{fig:overview}, we use a mask to remove the portrait from the latent space of the reference image.

During the training phase, we use a mask $\mathcal{M}$ to remove the portrait, compensating for the loss of identity and appearance information using the reference adapter. Additionally, we incorporate each frame’s portrait mask into the loss function, encouraging the model to focus more on the portrait region. The loss function is defined as follows:
\begin{equation}
	\mathcal{L}_p = \mathbb{E}\left[\|\left(1-\mathcal{M}\right)\left(\bm{\epsilon} -\bm{\epsilon}_\theta\left(\mathbf{z}_t,t,\mathbf{c}\right)\right) \|\right],
	\label{eq:Lp}
\end{equation}
where \(\bm{\epsilon} \sim \mathcal{N}(0,I) \in \mathbb{R}^{T\times h\times w\times c}\), and the portrait mask \(\mathcal{M} \in \left\{0,1\right\}^{T\times h\times w\times c}\). Finally, the total loss is formulated as:
\begin{equation}
	\mathcal{L} = \mathcal{L}_p + \mathcal{L}_{LDM}.
	\label{eq:Ltotal}
\end{equation}
\subsection{Lighting Controllable Portrait Animation}
\label{sec:lcpa}
In the relighting and animation stage, we incorporate the reference image, video fragment, and target lighting. When the portrait in the reference image corresponds to the same individual as that in the video fragment, we utilize DECA to extract the pose information from the video portrait and the shape information from the reference image, subsequently rendering a sequence of shading hints based on the lighting coefficients derived from the target lighting. However, in cases where the portraits in the video and reference image do not represent the same individual, we introduce a motion alignment module to mitigate the risk of identity leakage from the video portrait, which could compromise the quality of the generated output (for further details of motion alignment, please refer to the supplementary materials).

Following this, we input the shading hints and the reference image into the shading adapter and reference adapter. Given that the portrait in the reference image inherently contains its own lighting information, directly combining features may result in the original lighting dominating the shading hints, leading to ineffective relighting. To solve this, we adopt the concept of Composer \cite{huang2023composer}. This allows us to achieve portrait lighting manipulation by adjusting the direction of the lighting guidance within the set of conditions. The formula is as follows:
\begin{equation}
\hat{\bm{\epsilon}}_\theta(\mathbf{z}_t, \mathbf{c}) = \omega \left( \bm{\epsilon}_\theta(\mathbf{z}_t, \mathbf{c}_2) - \bm{\epsilon}_\theta(\mathbf{z}_t, \mathbf{c}_1) \right) +\bm{\epsilon}_\theta(\mathbf{z}_t, \mathbf{c}_1),
\label{eq:mccfg}
\end{equation}
here, \(\mathbf{c}_1\) and \(\mathbf{c}_2\) are two sets of conditions. If a condition exists in \(\mathbf{c}_2\) but not in \(\mathbf{c}_1\), its strength is enhanced by a weight \(\omega\). The larger \(\omega\), the stronger the condition. If a condition exists in both \(\mathbf{c}_1\) and \(\mathbf{c}_2\), \(\omega\) has no effect, and the condition strength defaults to 1.0.
In this way, we can set \(\mathbf{c}_2 = \mathbf{F}_s+ \mathbf{F}_r\) and \(\mathbf{c}_1 = \mathbf{F}_r\), where \(\mathbf{F}_s\) and \(\mathbf{F}_r\) are the features from the shading adapter and reference adapter. Since \(\mathbf{F}_s\) is present in \(\mathbf{c}_2\) but not in \(\mathbf{c}_1\), we can enhance the strength of the extrinsic feature by adjusting \(\omega\). At the same time, because both \(\mathbf{c}_1\) and \(\mathbf{c}_2\) contain \(\mathbf{F}_r\), the reference image’s portrait features remain intact. This allows us to achieve relightable portrait animation by utilizing classifier-free guidance for condition combination.

\section{Experiments}
\subsection{Implementation Details}
\textbf{Datasets.} We train our model on the CelebV-HQ \cite{zhu2022celebv} and VFHQ \cite{xie2022vfhq} datasets. Since the backbone of SVD \cite{blattmann2023stable} is sensitive to video quality, we first evaluate each video in two datasets with the video quality assessment method FasterVQA \cite{wu2023neighbourhood}, and remove videos with scores lower than 0.6. In the end, 37,644 videos remain for training. To ensure a fair comparison in experiments, we evaluate our method on the portrait video dataset HDTF \cite{zhang2021flow} and FFHQ \cite{karras2019style}.

\noindent\textbf{Training Details.} During the training phase, for the temporal attention layers of the SVD, we sample 16-frame video sequences to establish temporal consistency, with each frame at a resolution of $512\times512$. Unlike methods such as \cite{hu2024animate,ma2024follow}, which require two separate training stages, we update all the weights of both the SVD and two adapters simultaneously. The model is trained for 30,000 steps with a batch size of 8 using gradient accumulation, optimized by 8bit-Adam \cite{kingma2014adam} with a learning rate of $1\times10^{-5}$. 
\subsection{Metrics and Comparisons}
\textbf{Evaluation Metrics.} To evaluate the performance of our method, following \cite{cai2024real}, we relight the first 100 frames of each video in the HDTF dataset. Each video is rendered with four distinct lighting conditions derived from four different lighting-effect reference faces, resulting in a total of 44,000 frames for comprehensive comparison. Following \cite{nerffacelighting}, we use an off-the-shelf estimator \cite{feng2021learning} to calculate the Lighting Error (LE). Arcface \cite{deng2019arcface} is used to measure Identity Preservation (ID) between the relit results and the original images. To assess temporal consistency, we compute LPIPS \cite{zhang2018perceptual} between adjacent frames. We further employ an image quality assessment model \cite{pyiqa} and a video quality assessment model \cite{wu2023neighbourhood} to evaluate Image Quality (IQ) and Video Quality (VQ), respectively. Additionally, Fréchet Inception Distance (FID) \cite{heusel2017gans} and Fréchet Video Distance (FVD) \cite{skorokhodov2022stylegan} are used to measure video fidelity. In addition to objective evaluation, we conduct a user study in which 17 participants rate the videos based on three criteria: Lighting Accuracy (LA-User), Identity Similarity (ID-User), and Video Quality (VQ-User). Each criterion is rated on a scale of 1 to 5: poor, fair, average, good, and excellent. Finally, we calculate the average score for each criterion across participants.

\noindent\textbf{Comparative Methods. }For the portrait relighting task, we conduct a comparative analysis between LCVD and five state-of-the-art portrait relighting methods: DPR \cite{zhou2019deep}, SMFR \cite{hou2021towards}, NFL \cite{nerffacelighting}, StyleFlow \cite{10.1145/3447648}, and DiFaReli \cite{ponglertnapakorn2023difareli}, evaluating performance on both the HDTF and FFHQ datasets. For the portrait animation task, we compare LCVD with three state-of-the-art portrait animation methods: DaGAN \cite{hong2022depth}, StyleHEAT \cite{yin2022styleheat}, and AnimateAnyone \cite{hu2024animate}, using the HDTF dataset for evaluation.
\begin{table*}[t!]
    \centering
    \caption{Quantitative comparison of portrait relighting with DPR, SMFR, NFL, StyleFlow, and DiFaReli based on objective evaluation and user study on the HDTF video dataset. The best scores are highlighted in bold, and the second-best are underlined.}
    \vspace{-2mm}
    \label{tab:compare}
    \scalebox{1.0}
    {\begin{tabular}{cccccccc||ccc}
        \hline
        &\multicolumn{7}{c}{Objective Evaluation}&\multicolumn{3}{c}{User Study} \\
        \cmidrule(r){2-8} \cmidrule(r){9-11}
        Methods & LE$\downarrow$ & ID$\uparrow$ & LPIPS$\downarrow$ & IQ$\uparrow$ & VQ$\uparrow$ & FID$\downarrow$ & FVD$\downarrow$ & LA-User$\uparrow$ & ID-User$\uparrow$ & VQ-User$\uparrow$\\
        \hline\hline
        DPR \cite{zhou2019deep} & 0.768 & \textbf{0.730} & \underline{0.0295} & \underline{2.646} & 0.734 & \underline{44.57} & \underline{403.0} & \underline{3.423} & \underline{3.462} &\underline{3.125}\\
        SMFR \cite{hou2021towards} & \underline{0.747} & \underline{0.601} & 0.0333 & 1.057 & 0.588 & 60.50 & 551.6 & 3.047 & 2.877 & 2.604\\
        NFL \cite{nerffacelighting} & 0.784 & 0.199 & 0.0823 & 2.586 & \underline{0.766} & 96.17 & 819.3 & 2.894 & 2.553 & 2.398\\
        StyleFlow \cite{10.1145/3447648} & 0.932 & 0.474 & 0.1088 & 2.614 & 0.746 & 161.3 & 900.6 & 2.103 & 1.929 &1.563\\
        DiFaReli \cite{ponglertnapakorn2023difareli} & 0.783 & 0.531 & 0.1152 & 1.103 & 0.458 & 57.49 & 743.2 & 3.141 & 2.592 & 2.284\\
        \hline
        Ours & \textbf{0.738} & 0.585 & \textbf{0.0282} & \textbf{3.034} & \textbf{0.775} & \textbf{37.46} & \textbf{273.3} & \textbf{3.534} & \textbf{4.000} & \textbf{3.398}\\
        \hline\hline
    \end{tabular}}
    \vspace{-4mm}
\end{table*}

\begin{table}
    \centering
    \caption{Quantitative comparison of portrait relighting with NFL, StyleFlow and DiFaReli on the FFHQ dataset. The best scores are highlighted in bold, and the second-best are underlined.}
    \vspace{-2mm}
    \begin{tabular}{ccccc}
        \hline
        Methods & LE$\downarrow$ & ID$\uparrow$ & IQ$\uparrow$ & FID$\downarrow$\\
        \hline\hline
        NFL\cite{nerffacelighting} & \underline{0.892} & 0.253 & 3.020 & 118.9\\
        StyleFlow\cite{10.1145/3447648} & 1.042 & 0.485 & \underline{3.846} & 102.7\\
        DiFaReli\cite{ponglertnapakorn2023difareli} & \textbf{0.749} & \underline{0.687} & 1.591 & \textbf{25.98}\\
        \hline
        Ours & 0.938 & \textbf{0.765} & \textbf{4.465} & \underline{26.71}\\
        \hline
    \end{tabular}
    \vspace{-4mm}
    \label{tab:ffhq}
\end{table}
\begin{table}
    \centering
    \caption{Quantitative comparison of cross-identity portrait animation with DaGAN, StyleHEAT, and AnimateAnyone on the HDTF dataset. The best scores are highlighted in bold, and the second-best scores are underlined.}
    \vspace{-2mm}
    \scalebox{0.95}
    {\begin{tabular}{cccccc}
        \hline
        Methods & ID$\uparrow$ & POSE$\downarrow$ & IQ$\uparrow$ & VQ$\uparrow$ & FID$\downarrow$\\
        \hline\hline
        DaGAN\cite{hong2022depth} & 0.645 & \underline{3.935} & 1.005 & 0.528 & 107.4 \\
        StyHE.\cite{yin2022styleheat} & 0.201 & 34.58 & 1.554 & 0.612 & 149.9 \\
        AniAny.\cite{hu2024animate} & \underline{0.806} & 5.086 & \underline{2.744} & \underline{0.706} & \underline{69.85} \\
        \hline
        Ours & \textbf{0.876} & \textbf{3.805} & \textbf{3.021} & \textbf{0.717} & \textbf{49.11} \\
        \hline
    \end{tabular}}
    \vspace{-4mm}
    \label{tab:animate}
\end{table}
\subsection{Quantitative Evaluation}
In portrait video relighting, Table \ref{tab:compare} shows that our method outperforms other state-of-the-art methods in all metrics except for ID. Specifically, it improves video fidelity (FVD) by 32\%, image fidelity (FID) by 16\%, and image quality (IQ) by 14.6\% compared to the second-best method, demonstrating excellent video quality. While our method does not achieve the highest ID performance, this is because relighting in our method is applied during portrait animation, where ID information is derived only from the reference, unlike other methods that relight each frame individually. However, our method achieves the best ID performance in the user study, likely due to its higher-quality, more stable video synthesis, which visually aligns with better ID preservation. This also proves that the ID loss in our method is within an acceptable range for human perception.

Since NFL \cite{nerffacelighting}, StyleFlow \cite{10.1145/3447648}, and DiFaReli \cite{ponglertnapakorn2023difareli} are trained on the aligned FFHQ facial dataset, we compare our method on 500 FFHQ images for a fair evaluation. As shown in Table \ref{tab:ffhq}, our method outperforms the second-best method in identity preservation (ID) by 11.4\% and image quality (IQ) by 16.1\%. However, it does not achieve the best performance in lighting error (LE) and image fidelity (FID) because these methods are trained on FFHQ, while our model is trained on different video datasets, resulting in slightly lower lighting and fidelity performance. Notably, since our method is designed for video sequences and FFHQ is an image dataset, we replicate each image 16 times to form a video sequence in order to adapt the method for image testing.

In addition to portrait relighting, we use the lighting and shape from the reference image and the pose from the driving image to render shading hints, guiding our model to achieve cross-identity portrait animation, which we then evaluate. Beyond the previously mentioned metrics, we incorporate a POSE metric to assess the pose accuracy of the animated portraits, ensuring alignment with the poses in the driving video. The POSE evaluation method follows that of \cite{siarohin2021motion}, using a facial landmark detection model \cite{bulat2017far} to measure the pose error between the animated portraits and the driving portraits based on facial keypoints. As shown in Table \ref{tab:animate}, our method outperforms the other methods in all metrics, particularly achieving a 29.7\% improvement in image fidelity (FID), a 10.1\% improvement in image quality (IQ), and an 8.7\% improvement in identity preservation (ID) compared to the second-best method.
\begin{figure*}[!htbp]
	\centering
	\includegraphics[width=0.85\textwidth]{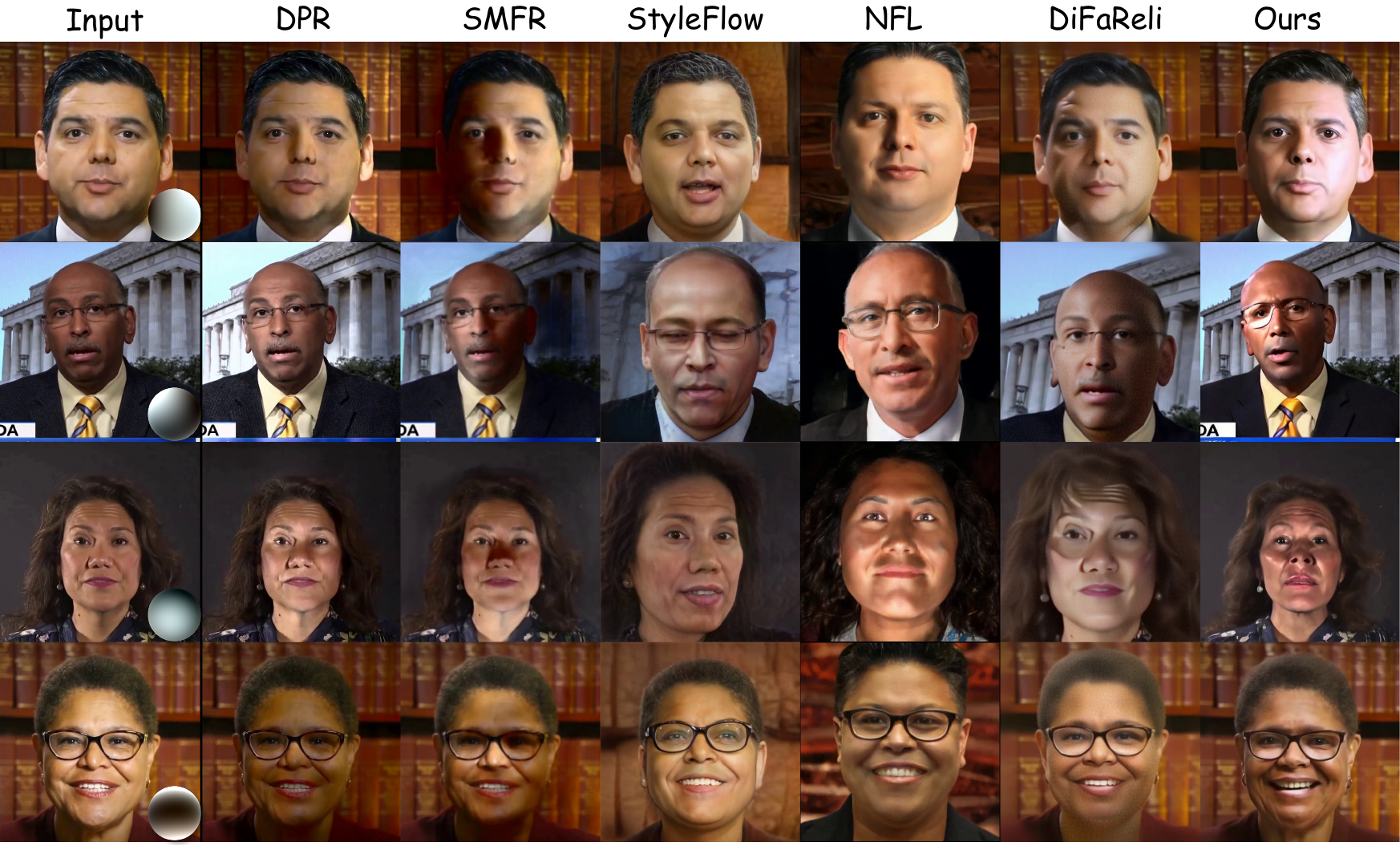}
	\caption{Qualitative comparisons with DPR \cite{zhou2019deep}, SMFR \cite{hou2021towards}, StyleFlow \cite{10.1145/3447648}, NFL \cite{nerffacelighting}, and DiFaReli \cite{ponglertnapakorn2023difareli}. The first column shows the input video frames, and the remaining columns present relighted results under various lighting conditions. Our method demonstrates more realistic performance, particularly in challenging cases such as side lighting.}
    \label{fig:compare_hdtf}
    \vspace{-0.18in}
\end{figure*}
\begin{figure}[!htbp]
	\centering
	\includegraphics[width=0.45\textwidth]{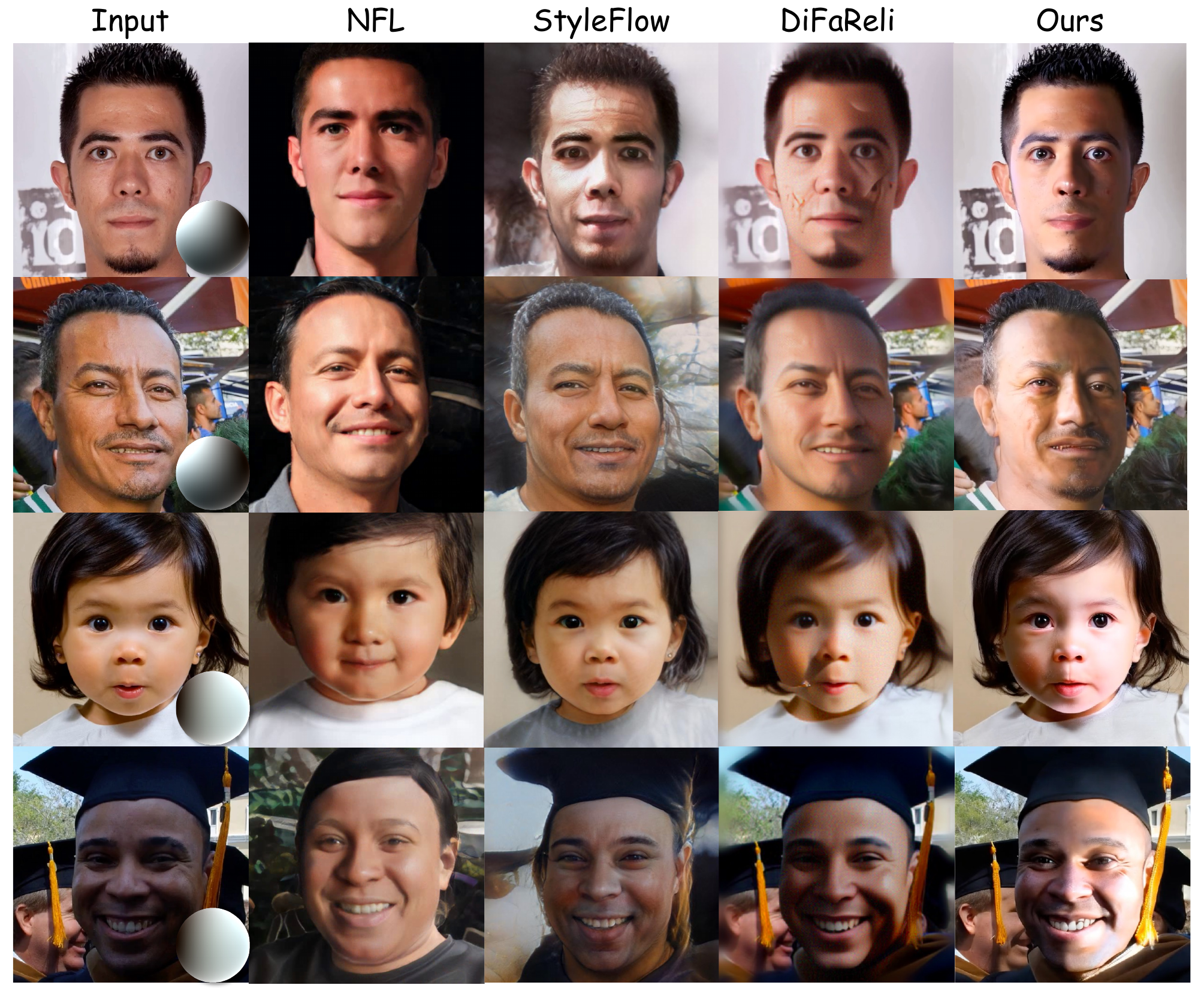}
	\caption{Qualitative comparison of portrait relighting with NFL \cite{nerffacelighting}, StyleFlow \cite{10.1145/3447648}, and DiFaReli \cite{ponglertnapakorn2023difareli} on the FFHQ dataset \cite{karras2019style}. The first column shows the input FFHQ portrait images, and the remaining column display the relighted results under various lighting conditions. Our method demonstrates more realistic results.}
    \label{fig:compare_ffhq}
    \vspace{-0.15in}
\end{figure}
\begin{figure}[!htbp]
	\centering
	\includegraphics[width=0.45\textwidth]{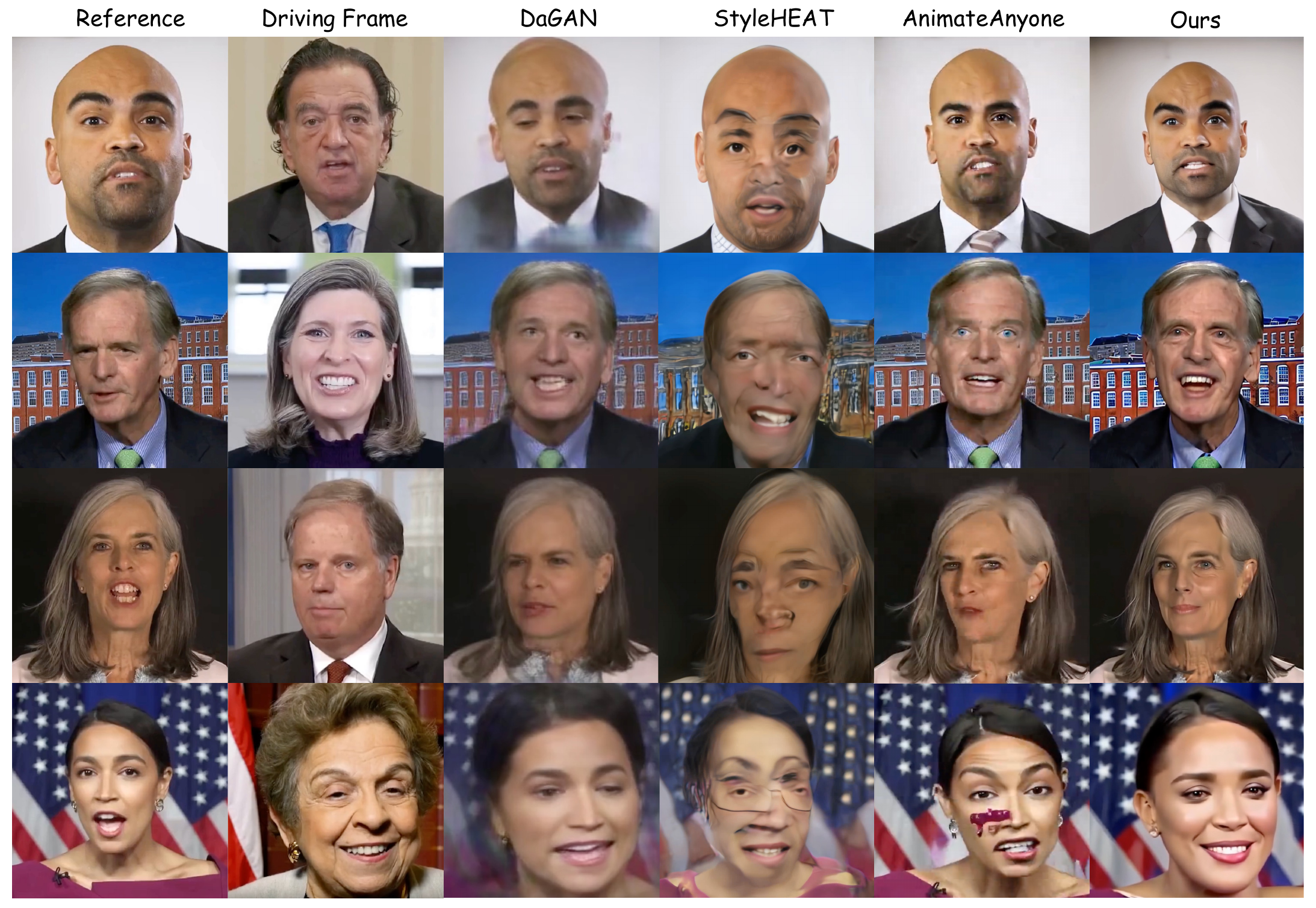}
	\caption{Qualitative comparison of cross-identity portrait animation with DaGAN \cite{hong2022depth}, StyleHEAT \cite{yin2022styleheat} and AnimateAnyone \cite{hu2024animate} on the HDTF dataset. Our method demonstrates more lifelike results.}
    \label{fig:compare_animate}
    \vspace{-0.18in}
\end{figure}

\begin{figure}[!htbp]
	\centering
	\includegraphics[width=0.45\textwidth]{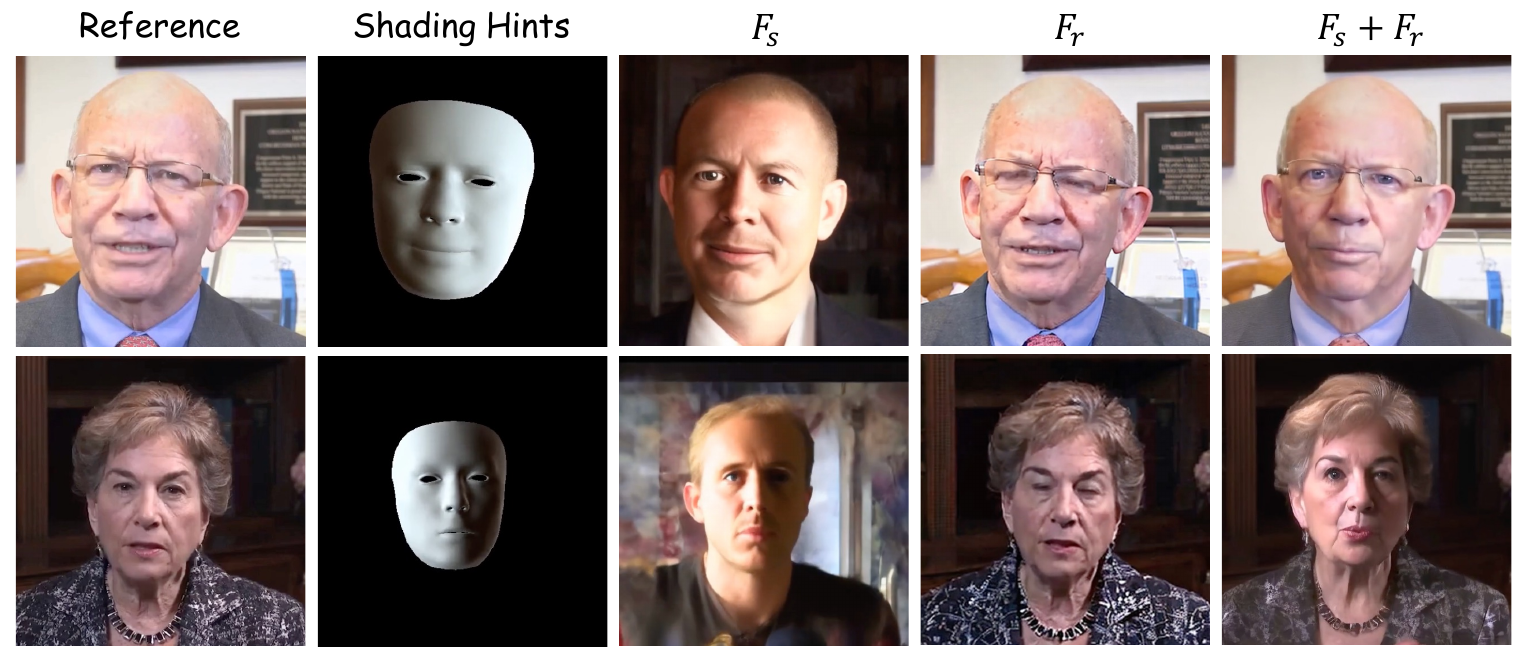}
	\caption{Ablation study comparing the performance of our model in portrait generation under different adapter combinations. $F_s$ represents using only the shading adapter, $F_r$ represents using only the reference adapter, and $F_s + F_r$ represents using both adapters together.}
    \label{fig:ablation_module}
    \vspace{-0.25in}
\end{figure}
\begin{figure}[!htbp]
	\centering
	\includegraphics[width=0.45\textwidth]{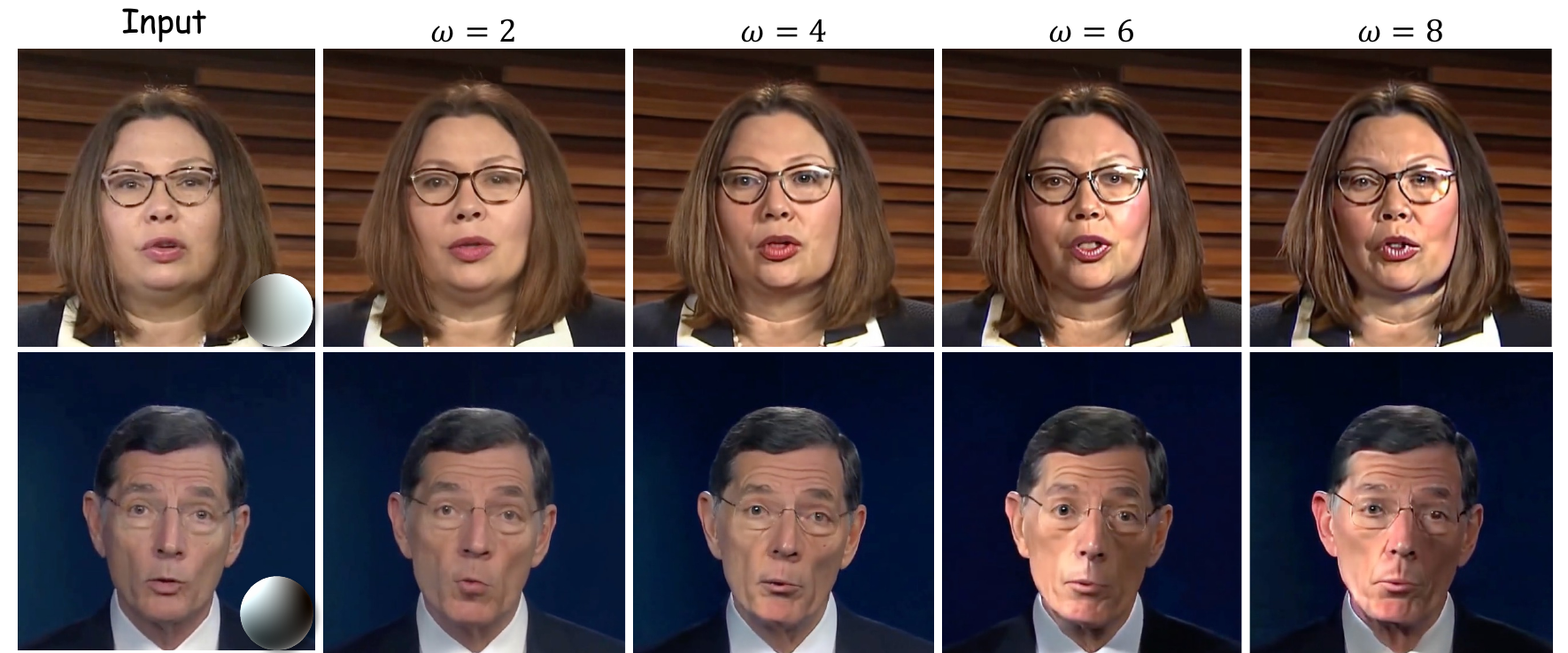}
	\caption{Ablation study comparing our model with varying strengths of multi-condition classifier-free guidance $\omega$. As $\omega$ increases, the relighting effect increasingly aligns with the target lighting; however, this comes at the cost of some loss of identity information and a decrease in image quality.}
    \label{fig:ablation}
    \vspace{-0.18in}
\end{figure}
\subsection{Qualitative Evaluation}
We compare our approach with previous portrait relighting methods on the HDTF dataset, including state-of-the-art face alignment-based approaches such as StyleFlow \cite{10.1145/3447648}, NFL \cite{nerffacelighting}, and DiFaReli \cite{ponglertnapakorn2023difareli}. Additionally, we compare our method with face alignment-free methods like DPR \cite{zhou2019deep} and SMFR \cite{hou2021towards}. The results are shown in Fig. \ref{fig:compare_hdtf}. We find that face alignment-based methods easily suffer from background detail loss and identity degradation, especially in pre-trained StyleGAN-based \cite{karras2020analyzing} methods like StyleFlow and NFL (e.g., see the results in the fourth and fifth columns, where the background details are completely lost, and the facial identity is inconsistent with the input). On the other hand, DiFaReli, based on a pre-trained diffusion model \cite{preechakul2021diffusion}, benefits from the DDIM inverse \cite{song2020denoising} method, which successfully reconstructs background details and preserves identity; however, it introduces noticeable artifacts on the face.

Although face alignment-free methods like DPR and SMFR achieve relighting without losing background and facial identity, the trade-off is a significant reduction in image quality, with the lighting appearing unnatural, as if a shadow has been cast over the image (e.g., in the first and second rows of the third column for SMFR). In contrast, our method in the final column greatly outperforms others in both image quality and the realism of the lighting effects. Notably, our approach accurately renders specular reflections on the face and eyes, as well as realistic shadows cast by facial muscles, while keeping identity loss within acceptable limits. The background details are also largely preserved. Overall, our approach demonstrates superior capability.

Since NFL, StyleFlow, and DiFaReli are trained on the aligned FFHQ dataset, we visualize the relighting results on FFHQ for a fair comparison. As shown in Fig. \ref{fig:compare_ffhq}, NFL and StyleFlow lose background details and alter the portrait identity. DiFaReli preserves background details but introduces facial artifacts, lowering image quality. In contrast, our method maintains background details and identity consistency, achieving optimal image quality.

Additionally, we compare our method with DaGAN, StyleHEAT, and AnimateAnyone for portrait animation. As shown in Fig. \ref{fig:compare_animate}, while DaGAN preserves the pose from the driving frame, the portrait identity differs significantly from the reference, and the image quality is low. StyleHEAT introduces distortions in cross-identity portrait animation, and although AnimateAnyone, a diffusion model guided by a reference-net, generates higher image quality, it still suffers from identity loss and occasional facial artifacts.
\subsection{Ablation Study}
\textbf{Effectiveness of Adapters.} Our method constructs intrinsic and extrinsic feature subspaces using the reference and shading adapters, respectively, enabling relightable portrait animation by merging these subspaces. We conduct an ablation study with different adapter combinations. First, when retaining only the shading adapter as shown in Fig. \ref{fig:ablation_module}, the column labeled $F_s$  illustrates that the generated portrait’s pose and lighting align with the shading hints, indicating that only the extrinsic features are transferred. When only the reference adapter is used, the column labeled $F_r$ shows that the generated portrait closely resembles the reference with only minor variations, such as blinking, indicating intrinsic feature preservation. When both adapters are used, the column labeled $F_s + F_r$ demonstrates that the generated portrait not only matches the pose and lighting of the shading hints but also maintains the identity and appearance of the reference.

\noindent\textbf{Effectiveness of Guidance Strength.} In Fig. \ref{fig:ablation}, we visualize the relighting results for different $\omega$ values. When $\omega = 2$, the lighting effect is minimal, with only small differences from the input image, resulting in good identity retention. In contrast, when $\omega = 8$, the lighting effect closely aligns with the target lighting, but this also leads to reduced image quality and some loss of identity retention. The primary reason for this phenomenon is that as $\omega$ increases, the proportion of extrinsic features grows, while the proportion of intrinsic features diminishes, resulting in a degradation of identity information from the reference image. Consequently, higher values of $\omega$ enhance lighting effects but lead to greater identity loss.

\section{Conclusion}
In this paper, we introduce the Lighting Controllable Video Diffusion model (LCVD) for high-fidelity, relightable portrait animation. By distinguishing between intrinsic and extrinsic facial features, our approach effectively preserves identity and appearance while enabling precise control over lighting and pose. We propose a novel framework that leverages reference and shading adapters to construct feature subspaces and incorporates multi-condition classifier-free guidance to fine-tune the lighting effects. Our extensive experimental results demonstrate that LCVD outperforms existing methods, providing significant improvements in lighting realism, image quality, and video consistency. 
{
    \small
    \bibliographystyle{ieeenat_fullname}
    \bibliography{main}
}

 \clearpage
\setcounter{page}{1}
\maketitlesupplementary

\section{Ablation Study}
\textbf{Effectiveness of Adapters. }The shading adapter maps shading hints to the extrinsic feature subspace, while the reference adapter maps the reference to the intrinsic feature subspace. The combination of features from these different subspaces enables various effects, such as controlling lighting magnitude, maintaining identity, and enhancing image generation quality. To investigate the effectiveness of these adapters, we conducted an ablation study with different adapter combinations.

First, we retain only the reference adapter, as shown in Table \ref{tab:ablation_adapters} under the row $F_r$. In this case, the lighting error is significant (LE is large), while identity preservation is excellent (ID is high). This indicates that the model preserves intrinsic features well but fails to capture extrinsic features. Conversely, when we retain only the shading adapter, as shown in the row $F_s$, the lighting error is minimal (LE is small), but identity preservation is almost nonexistent (ID approaches 0). This suggests that the model transfers extrinsic features effectively while neglecting intrinsic features.

When both adapters are retained, we observe significant improvements in intrinsic feature preservation compared to using only the shading adapter and significant improvements in extrinsic feature transfer compared to using only the reference adapter. Moreover, the image quality also achieves its optimal level under this configuration.

\noindent\textbf{Effectiveness of Guidance Strength.} This method utilizes a multi-condition classifier-free guidance approach to control the lighting magnitude through the classifier-free guidance mechanism \cite{ho2022classifier}. The strength of the guidance, represented by $\omega$, directly affects the lighting intensity.

To evaluate the impact of $\omega$, we conduct an ablation study with varying values, as shown in Table \ref{tab:ablation_omega}. As $\omega$ increases, the lighting effect improves (LE decreases), but identity preservation deteriorates (ID decreases). Notably, image quality reaches its peak at $\omega=4$. However, setting $\omega$ too high can lead to a decline in image quality. Therefore, lighting effects, identity preservation, and image quality can be balanced by appropriately adjusting the value of $\omega$.
\begin{table}
    \centering
    \caption{Quantitative comparison of ablation study with different adapter combinations on the HDTF dataset. $F_r$ denotes using only the reference adapter, $F_s$ denotes using only the shading adapter, and $F_s+F_r$ represents using both adapters. The best scores are highlighted in bold, and the second-best are underlined.}
    \vspace{-2mm}
    \begin{tabular}{ccccc}
        \hline
        Methods & LE$\downarrow$ & ID$\uparrow$ & IQ$\uparrow$ & FID$\downarrow$\\
        \hline\hline
        $F_r$ & 1.071 & \textbf{0.802} & \underline{1.662} & \textbf{35.61}\\
        $F_s$ & \textbf{0.582} & 0.028 & 1.248 & 56.63\\
        \hline
        $F_s+F_r$ & \underline{0.738} & \underline{0.585} & \textbf{3.034} & \underline{37.46}\\
        \hline
    \end{tabular}
    \label{tab:ablation_adapters}
\end{table}
\begin{table}
    \centering
    \caption{Quantitative comparison of the ablation study on the impact of different guidance strengths $\omega$ on lighting (LE), identity (ID), and image quality (IQ) on the HDTF dataset. From left to right, each metric is shown as it changes with increasing $\omega$. The best scores are highlighted in bold, and the second-best are underlined.}
    \vspace{-2mm}
    \begin{tabular}{ccccc}
        \hline
        Methods & $\omega=2$ & $\omega=4$ & $\omega=6$ & $\omega=8$\\
        \hline\hline
        LE$\downarrow$ & 1.079 & 0.809 & \underline{0.744} & \textbf{0.681}\\
        ID$\uparrow$ & \textbf{0.728} & \underline{0.603} & 0.563 & 0.503\\
        IQ$\uparrow$ & 2.611 & \textbf{2.988} & \underline{2.954} & 
        2.856\\
        \hline
    \end{tabular}
    \vspace{-4mm}
    \label{tab:ablation_omega}
\end{table}
\section{Motion Alignment}
\label{sec:rationale}
As shown in Fig. \ref{fig:overview}, during the relighting and animation stages, we use a video to animate the reference image, ensuring that the lighting effect of the relit portrait is consistent with that of the target lighting. In the inference stage, since the portrait in the video and the reference image come from different identities, directly using the shading hints of the portrait from the video to animate the reference image would cause the generated portrait to resemble the one from the driving video. This leads to identity leakage during animation, degrading the animation quality. We propose two motion alignment methods: (1) a relative displacement-based motion alignment method and (2) a portrait scale consistency-based motion alignment method.

\noindent\textbf{Relative Displacement-based Motion Alignment.} This motion alignment method is designed to use the reference image as the first frame, with subsequent motions based on this initial frame. The motion guidance for the reference frame is achieved by leveraging the relative displacement between consecutive frames in the driving video. First, we use DECA to extract the pose sequence \(\mathbf{P} = \{p^v_1, p^v_2, \dots, p^v_n\}\) and the expression sequence \(\mathbf{E} = \{e^v_1, e^v_2, \dots, e^v_n\}\) from each frame of the driving video, along with the pose \(p^R\) and shape \(s^R\) from the reference image. Next, we calculate the relative pose offsets \(\bm{\Delta} \mathbf{P} = \{0, p^v_2 - p^v_1, \dots, p^v_n - p^v_1\}\) for each frame with respect to the first frame. Using the reference image’s pose \(p^R\) as the base pose, we then apply these relative offsets to obtain an aligned pose sequence \(\mathbf{P}^{align} = \{p^R, p^R + (p^v_2 - p^v_1), \dots, p^R + (p^v_n - p^v_1)\}\). Finally, we combine the expression sequence \(\mathbf{E}\) with the reference image’s shape \(s^R\) and the aligned pose sequence \(\mathbf{P}^{align}\). These parameters are then input into Eq. \ref{eq:flame} to obtain \(\mathbf{FLAME}(s^R, \mathbf{P}^{align}, \mathbf{E})\), which, along with the spherical harmonic lighting coefficients \(l\) from the target lighting, is used to render the shading hints for each frame.

\noindent\textbf{Portrait Scale Consistency-based Motion Alignment.} The relative displacement-based alignment method relies on using the reference image as the base frame for relative motion. However, this approach does not ensure perfect spatial alignment between the pose of the generated portrait and the driving video. To address this, we propose an alternative motion alignment method aimed at achieving perfect alignment between the generated portrait's pose and that of the driving video. Specifically, we first use DECA to extract the pose sequence \(\mathbf{P} = \{p^v_1, p^v_2, \dots, p^v_n\}\) and the expression sequence \(\mathbf{E} = \{e^v_1, e^v_2, \dots, e^v_n\}\) from each frame of the driving video, along with the shape \(s^R\) from the reference image. These parameters are then input into Eq. \ref{eq:flame} to compute \(\mathbf{FLAME}(s^R, \mathbf{P}, \mathbf{E})\). Combined with the spherical harmonic lighting coefficients \(l\) from the target lighting, this process renders the shading hints for each frame.
\section{Shading and Reference Adapter Network Architecture}
\begin{figure}[!htbp]
	\centering
	\includegraphics[width=0.5\textwidth]{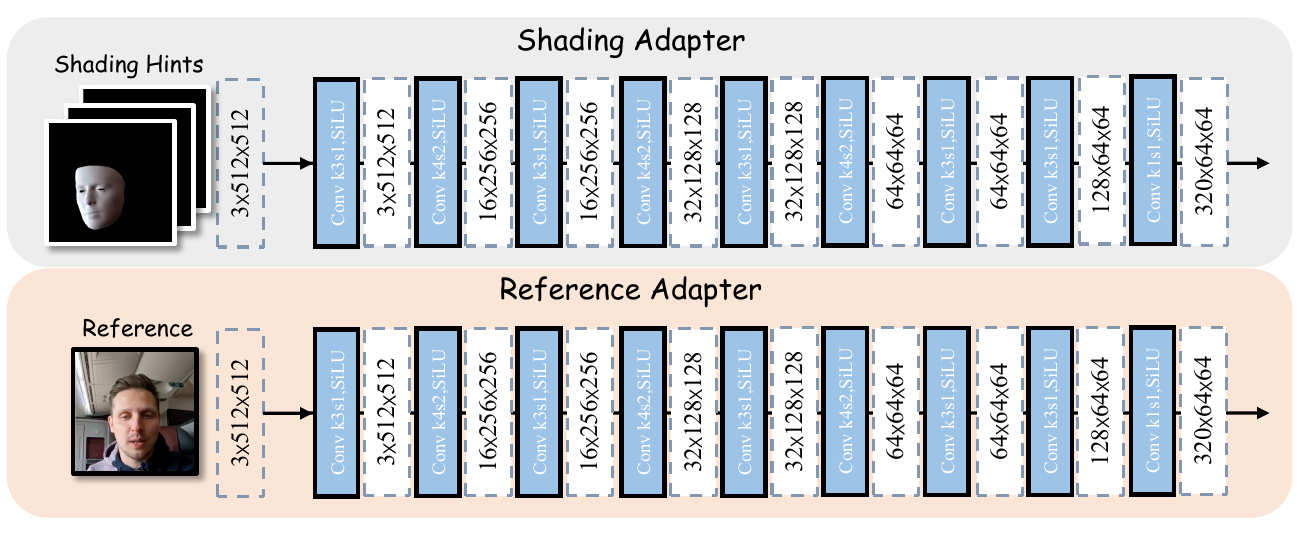}
        \caption{Network architecture of the shading adapter and reference adapter, where $k$ denotes the kernel size and $s$ denotes the stride. These two networks have the same structure but do not share weights and are updated alongside SVD during the training phase.}
    \label{fig:adapter_network}
    \vspace{-0.25in}
\end{figure}

As shown in Fig. \ref{fig:adapter_network}, the network architecture of the shading adapter and reference adapter is illustrated. These two networks map shading hints and the reference image into the \textit{extrinsic feature subspace} and \textit{intrinsic feature subspace} of SVD’s feature space, respectively. As depicted in Fig. \ref{fig:overview}, the two features are fused with the features from the first convolutional layer of SVD. Therefore, the shading hints and reference image must match the spatial dimensions and channel count of the output from SVD’s first convolutional layer. To achieve this, we designed the network structure shown in Fig. \ref{fig:adapter_network}.

Moreover, since SVD is designed for video sequence generation, the output dimensions of its first layer include an additional temporal dimension \(F\), resulting in an output shape of \(B \times F \times C \times H \times W\). Accordingly, the input to the shading adapter is a sequence of shading hints with dimensions \(B \times F \times C \times H \times W\). For the reference image, which consists of a single frame with dimensions \(B \times 1 \times C \times H \times W\), we duplicate the reference \(F\) times to obtain dimensions \(B \times F \times C \times H \times W\) before feeding it into the reference adapter.
\section{Long Video Sequence Generation}
Since our model is based on the SVD backbone, which is limited to generating video sequences of 16 frames at a time, we tackle the challenge of animating portrait videos of arbitrary length by utilizing the diffusion model sampling method proposed in \cite{zhang2024mimicmotion}. To ensure smooth transitions between consecutive video segments, we implement a 6-frame overlap strategy. In our experiments, we employ DDIM with 25 sampling steps and set the default guidance weight $\omega$ to 4.5. For a 100-frame video, this method takes approximately two minutes and 10 GB of VRAM to perform inference on an NVIDIA 4090 GPU.

\end{document}